\documentclass[11pt]{article}

\usepackage[preprint]{acl}

\usepackage{times}
\usepackage{latexsym}

\usepackage[T1]{fontenc}
\usepackage[utf8]{inputenc}

\usepackage{microtype}

\usepackage{inconsolata}

\usepackage{graphicx}

\usepackage[capitalize]{cleveref}

\usepackage{tcolorbox}

\usepackage{booktabs}
\usepackage{multirow}

\usepackage{tabularx}
\usepackage{makecell}
\usepackage{ragged2e}
\usepackage{array}

\title{Evaluating the Diagnostic Robustness of Vision-Language Models Under Visual and Textual Perturbations}

\author{
  \textbf{Ali Khoramfar}\textsuperscript{1},
  \textbf{Mohammad Javad Dousti}\textsuperscript{1},
  \textbf{Alireza Mohamadian}\textsuperscript{2},
  \textbf{Heshaam Faili}\textsuperscript{1} \\
  \textsuperscript{1}Department of Electrical and Computer Engineering, College of Engineering, \\
  University of Tehran, Tehran, Iran \\
  \textsuperscript{2}Advanced Diagnostic and Interventional Radiology Research Center (ADIR), \\
  Tehran University of Medical Sciences, Tehran, Iran \\
  \texttt{\textsuperscript{1}\{khoramfar, mjdousti, hfaili\}@ut.ac.ir}
, \texttt{\textsuperscript{2}alirezamohamadian.md@gmail.com}
}

\begin{document}
\maketitle
\begin{abstract} 
Standard accuracy metrics for VLMs often mask significant reliability failures in sensitive domains. In this work, we utilize a histopathology-confirmed brain MRI dataset to systematically assess the diagnostic robustness of four VLM families under controlled visual and textual perturbations. By reordering anatomical slices and swapping target label positions, we evaluate prediction stability under controlled presentation changes. Our results reveal significant vulnerabilities in presentation-order stability, with models exhibiting prediction flips in up to 48.9\% of cases under simple sequence reversals. We further identify a strong sensitivity to textual ordering, where label reordering triggers inconsistent diagnoses in up to 67.8\% of cases despite identical visual inputs. Evidence-ablation tests further reveal diagnostic overcommitment: models generate categorical diagnoses in up to 76.1\% of cases after expert-annotated lesion slices are removed. These results show that aggregate accuracy can mask diagnostic instability under sequential presentation and textual framing. Our findings highlight the need for stability-based evaluation before VLMs are considered for safety-critical clinical applications. 
\end{abstract}

\section{Introduction}
\label{sec:intro}

Vision-Language Models (VLMs) have emerged as a dominant paradigm in Artificial Intelligence (AI) by bridging the long-standing gap between visual perception and natural language understanding within a unified architecture \citep{zhang_vision-language_2024}. Through large-scale pre-training on massive multimodal corpora, these models demonstrate highly advanced cross-modal reasoning capabilities across a wide range of complex downstream tasks \citep{wu_multimodal_2023, yin_survey_2024}.

This success has accelerated the exploration of VLMs in high-stakes and specialized domains \citep{liu2024few}. In healthcare, multimodal foundation models and domain-specific adaptations such as MedGemma \citep{sellergren2026medgemma15technicalreport} have become a major focus of active research. Current studies extensively explore their potential for complex clinical workflows, ranging from medical visual question answering to automated radiology report generation \citep{hartsock_vision-language_2024}.

Despite this surge in research interest, commonly used evaluation protocols remain incomplete for assessing clinical reliability. The Natural Language Processing (NLP) and medical AI communities traditionally rely on aggregate performance on static benchmarks to measure clinical utility. However, recent critical analyses suggest that standard evaluations often fail to measure the actual underlying skills they intend to assess \citep{alaaposition}. High success rates on curated multiple-choice questions do not necessarily guarantee that a model possesses the robust reasoning abilities required for real-world clinical applications.

Studies on visual cognition reveal a disconnect in model behavior. While VLMs can generate startlingly human-like responses, they frequently fail at basic causal and physical reasoning tasks \citep{schulze2025visual}. This raises a critical safety concern regarding whether these models' decisions remain reliably anchored to task-relevant visual evidence or are unduly influenced by statistical patterns in the input presentation.

This disconnect highlights a pressing need to broaden the evaluation paradigm toward diagnostic robustness and prediction consistency. Ideally, clinically equivalent presentations of the same available evidence should yield stable patient-level diagnoses. However, VLM predictions can change substantially under controlled presentation variations \citep{Michalkiewicz_2025_ICCV}. Assessing these visual sensitivities is therefore important for evaluating prediction stability in safety-critical clinical settings.

In this work, we systematically evaluate whether state-of-the-art VLMs maintain consistent diagnoses under controlled changes in visual or textual presentation. We further examine the consequences of prediction changes and evaluate whether sensitivity persists on an external cohort.

To study this question in a setting that allows precise spatial control over visual information, we instantiate our study on structural brain MRI data. Multi-slice medical imaging yields a highly structured testbed for such evaluations. Because these scans consist of sequential cross-sections, the clinical evidence is distributed across space, allowing us to precisely localize, reorder, or withhold specific visual cues to isolate their direct impact on model decisions.

Specifically, we pair clinical scans with expert-defined region annotations that pinpoint the slices containing diagnostic lesions. These annotations enable us to construct controlled input sequences that reorder, reposition, or omit the primary clinical evidence. By evaluating model behavior under these structured variations, we measure the extent to which diagnostic decisions remain stable across changes in visual and textual presentation.

Our main contributions are summarized as follows:
\vspace{-0.1cm}
\begin{itemize}

    \item \textbf{Proposed a perturbation-based framework for diagnostic robustness:}  
    We present a paired evaluation framework that measures prediction stability under controlled visual and textual presentation modifications.
    
    \item \textbf{Demonstrated sensitivity to visual and textual presentation:}  
    We show that sensitivity to slice sequencing and label ordering occurs across generalist and medically fine-tuned models, and extends to an external brain MRI task.
    
    \item \textbf{Characterized diagnostic overcommitment under targeted evidence ablation:} We analyze prediction changes and show that models may remain categorical after expert-annotated lesion evidence is removed.
\end{itemize}

The rest of this paper is structured as follows. \cref{sec:related_work} reviews relevant literature, while \cref{sec:methodology} describes our evaluation data and experimental setup. Finally, \cref{sec:results} discusses our findings, and \cref{sec:conclusion} provides our conclusions.

\begin{figure*}[t!]
  \centering
  \includegraphics[width=0.90\textwidth]{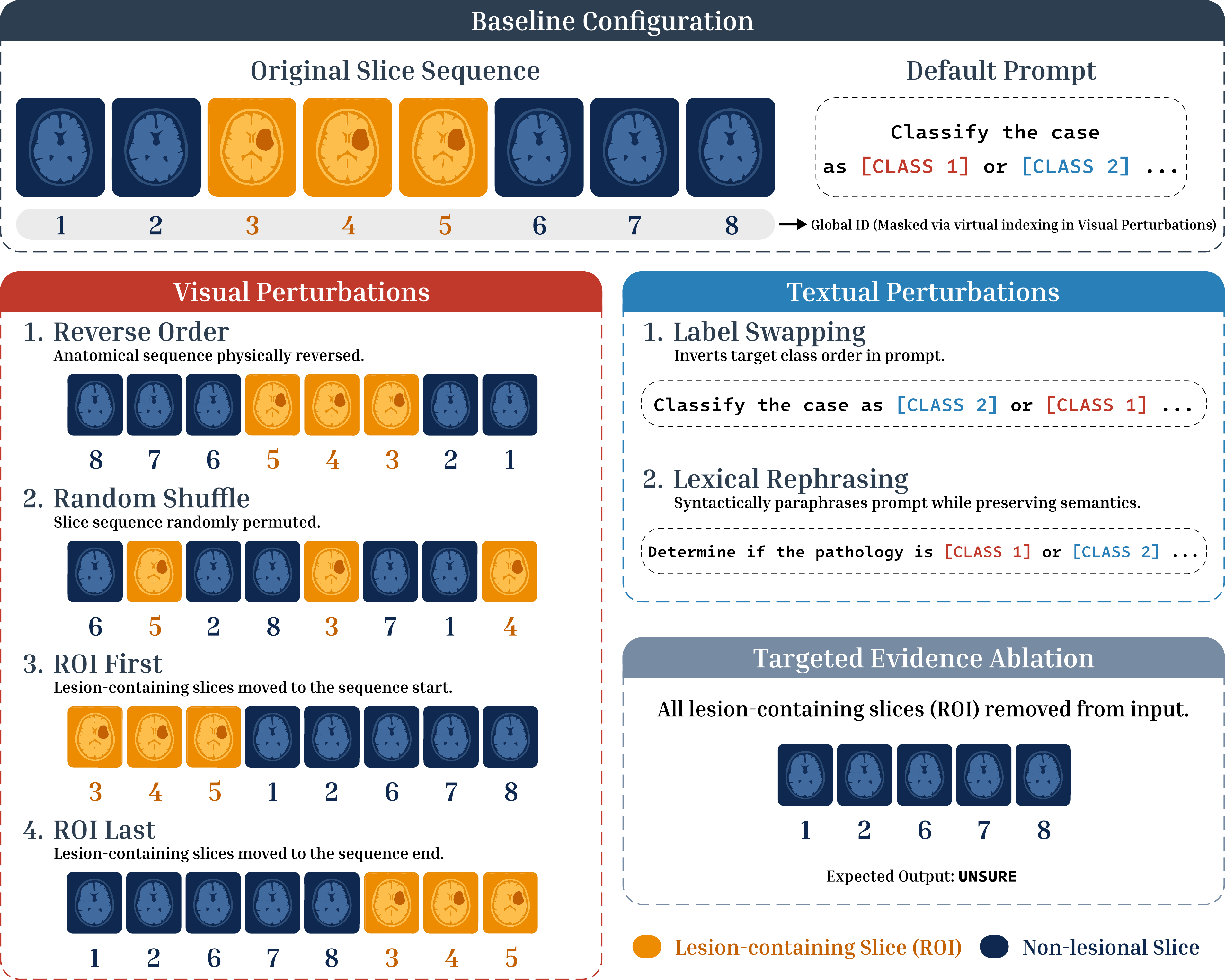}
  \caption{Overview of the evaluation framework, illustrating visual perturbations, textual perturbations, and targeted evidence ablation configurations. Slice indices represent the original sequence order for illustration.}
  \label{fig:main_workflow}
\end{figure*}

\section{Related Work}
\label{sec:related_work}

To better understand the reliability of multimodal models, recent studies have increasingly shifted focus toward diagnostic robustness and prediction consistency. \citet{rosenfeld_2025} demonstrated that model outputs fluctuate significantly under benign, meaning-preserving variations. Similarly, \citet{chou_mmr3_2025} empirically showed that models frequently yield contradictory responses to semantically equivalent queries, establishing that prediction consistency operates as a distinct dimension of reliability.

In exploring these vulnerabilities, a substantial body of literature has investigated cross-modal adversarial attacks \citep{babu_coordinated_2025}. These studies typically introduce coordinated noise to disrupt the alignment between vision and language. While these adversarial approaches are highly valuable for testing worst-case security boundaries, clinical environments also require evaluating models under natural variations and evidence-preserving visual perturbations.

A key element of this natural variation involves how models process spatial and sequential structures. \citet{Michalkiewicz_2025_ICCV} highlighted that foundation models exhibit viewpoint instability and struggle with spatial disentanglement. This spatial sensitivity is particularly relevant to medical imaging, where diagnostic evidence is distributed across volumetric planes, raising questions about whether model predictions remain stable across presentation order.

Beyond visual sensitivities, researchers have identified a disproportionate reliance on linguistic priors in multimodal reasoning. \citet{10.1007/978-3-032-06004-4_32} diagnosed textual biases in clinical AI. By swapping images or text between opposing diagnostic labels, they demonstrated that strong textual signals often overshadow image-based pathologies, causing models to prioritize text over visual cues.

This reliance on language frequently compromises the faithfulness of the models' internal reasoning. \citet{moll_medical_faithfulness_2025} reported that answer accuracy and explanation quality are often decoupled in medical models. Their findings indicate that models tend to follow injected textual cues rather than demonstrate true visual grounding, highlighting a marked sensitivity to textual framing.

Recent clinical evaluations provide a theoretical basis for these observed behaviors. \citet{han_decoupling_2025} explicitly separated visual parsing, defined as identifying findings on radiologic images, from diagnostic reasoning, which involves translating those findings into clinical diagnoses. Their analysis revealed that while current models possess strong text-based diagnostic logic, their actual visual parsing capabilities remain the principal bottleneck constraining performance.

Recent medical benchmarks have shifted toward progressive, multi-stage clinical workflows. For instance, \citet{xu2025medatlas} proposed a multi-round diagnostic framework to evaluate error propagation across sequential examinations. In the specific domain of neuro-oncology, \citet{guo2026mmneuroonco} developed a brain tumor MRI benchmark incorporating structured semantic attributes evaluations. Although these benchmarks expand the scope of medical VLM evaluation, they primarily assess diagnostic accuracy under static, canonical presentation formats.

To address these limitations, recent benchmarks evaluate broader trustworthiness dimensions such as OOD noise and safety \citep{NEURIPS2024_fde7f40f}, or reasoning faithfulness on 2D radiographs \citep{moll_medical_faithfulness_2025}. While these studies expand medical VLM evaluation, they primarily assess static 2D inputs or text-only rationales. Consequently, there remains a gap in understanding whether diagnostic decisions on volumetric multi-slice data remain stable under controlled visual and textual presentation changes. We address these concerns by systematically evaluating VLM predictions under visual and textual perturbations, alongside targeted evidence ablation.

\section{Methodology}
\label{sec:methodology}

To evaluate the diagnostic robustness of VLMs, we designed an evaluation pipeline that subjects the models to controlled visual and textual presentation perturbations, as schematically illustrated in \cref{fig:main_workflow}. This setup aims to isolate the impact of spatial presentation and textual framing, minimizing confounding variables to better observe model consistency under varying input conditions.

\subsection{Dataset and Clinical Task}
\label{sec:methodology:dataset}

The evaluation relies on a rigorously curated brain MRI dataset comprising 4,091 axial slice images. These scans were retrospectively acquired from 90 human subjects across multiple academic hospitals. Utilizing this specific clinical cohort---distinct from massively scraped public benchmarks---helps mitigate the risk of train-test data contamination.

The cohort is strictly balanced, consisting of 45 Glioblastoma (\texttt{GBM}) and 45 Brain Metastasis (\texttt{MET}) cases acquired across four academic hospitals using 1.5T Siemens MRI systems. For each subject, we evaluate two routinely acquired volumetric MRI sequences: T1-weighted contrast-enhanced (T1CE) and T2-weighted (T2). Extended clinical definitions and sequence statistics are provided in \cref{sec:appendix:dataset}.

We selected this specific differential diagnosis because it represents a clinically challenging task with high visual similarity between classes \citep{Cha1078}. This intrinsic difficulty provides a rigorous testbed to probe model consistency near complex decision boundaries. Furthermore, to ensure a reliable diagnostic reference, all ground-truth labels are histopathology-confirmed, derived directly from post-surgical laboratory pathology reports rather than initial radiological interpretations.

The primary task requires the model to perform a sequence-level differential diagnosis (\texttt{GBM} vs. \texttt{MET}) based entirely on the provided MRI slices. The models are explicitly instructed to output \texttt{UNSURE} if the visual findings are ambiguous or insufficient. To facilitate our perturbation framework, two board-certified neuroradiologists provided pixel-level consensus masks, allowing us to identify which slices intersect the annotated lesion region of interest (ROI) and which do not (detailed in \cref{sec:appendix:annotation}).

\subsection{Evaluated Models}
\label{sec:methodology:models}

To cover diverse VLM configurations, we evaluate four state-of-the-art models spanning different architectural and training paradigms:
\begin{itemize}
    \item \textbf{GPT-5.4} \citep{openai2025gpt5} \& \textbf{\mbox{Gemini-2.5-Pro}} \citep{comanici2025gemini}: Selected as leading proprietary generalists with extensive multimodal context windows and advanced reasoning capabilities.
    \item \textbf{Qwen3-VL-32B} \citep{bai2025qwen3vltechnicalreport}: Included as a highly capable open-source generalist to investigate whether accessible, open-source training paradigms exhibit similar structural sensitivities.
    \item \textbf{MedGemma-27B} \citep{sellergren2026medgemma15technicalreport}: Selected as a domain-specific, medically fine-tuned foundation model. Evaluating this model tests whether medical adaptation mitigates sensitivity to presentation changes.
\end{itemize}
Specific API snapshots and details regarding our handling of model constraints for long sequences are documented in \cref{sec:appendix:setup}.

\paragraph{Supervised Medical Imaging References.} To establish a non-VLM reference, we evaluate traditional radiomics classifiers---Support Vector Machines (SVM) and Random Forest (RF)---trained on expert-delineated 3D ROI features using patient-level 5-fold cross-validation.

\subsection{Cross-Task External Cohort}
\label{sec:methodology:external}

For cross-task external evaluation, we evaluate 100 subjects from the multi-institutional BraTS dataset \citep{6975210}. The cohort contains 50 high-grade glioma (\texttt{HGG}) and 50 low-grade glioma (\texttt{LGG}) cases, with one T1CE and one T2 volume per subject. We apply the same virtual-indexing protocol and evaluate sequence reversal and target-label reordering using GPT-5.4 and Qwen3-VL-32B as representative proprietary and open-source generalist models. Additional cohort selection and preprocessing details are provided in \cref{sec:appendix:external_cohort}.

\subsection{Evaluation Pipeline and Virtual Indexing}
\label{sec:methodology:pipeline}

To prevent models from exploiting slice filenames or embedded metadata, we implement a \textit{Virtual Indexing Mechanism}. Original file metadata is stripped, and each slice is presented with a dynamically generated neutral tag (e.g., ``\texttt{Image 1:}'', ``\texttt{Image 2:}'').

An internal presentation-to-global mapping allows model-reported evidence indices to be mapped back to anatomical slice IDs without exposing those IDs to the model. All queries use \texttt{temperature = 0} to minimize sampling-induced variability. The complete baseline prompt is provided in \cref{sec:appendix:prompt}.

\subsection{The Perturbation Framework}
\label{sec:methodology:perturbations}

Our experimental engine evaluates each human subject under a baseline condition and multiple distinct perturbation configurations, categorized as follows:

\paragraph{Baseline Configuration.}
Slices are provided in their natural anatomical sequence as acquired by the MRI scanner. The standard clinical prompt is utilized, serving as the reference diagnostic decision for each subject.

\paragraph{Visual Perturbations.}
The textual prompt remains fixed while image presentation order is modified:
\begin{itemize}
    \item \textbf{Reverse Order:} The anatomical slice sequence is strictly reversed, preserving slice content and local adjacency while changing presentation direction.
    \item \textbf{Random Shuffle:} Slices are reordered using deterministic random permutations. This preserves available slices but disrupts inter-slice continuity, serving as a stress test of evidence aggregation. We evaluate three independent deterministic permutations.
    \item \textbf{ROI First \& ROI Last:} Slices intersecting the expert-annotated lesion are moved to the beginning or end of the input sequence, respectively, while preserving the relative order within the ROI and non-ROI groups.
\end{itemize}

\paragraph{Textual Perturbations.}
The visual sequence remains unchanged while textual presentation is modified:
\begin{itemize}
    \item \textbf{Label Swapping:} Target-class order is reversed from \texttt{GBM/MET} to \texttt{MET/GBM} in the instruction and output schema; images remain identical.
    \item \textbf{Lexical Rephrasing:} The baseline prompt is paraphrased while preserving the clinical task, target labels, and output schema.
\end{itemize}

These configurations target distinct properties: Reverse Order and the textual variants evaluate presentation stability, ROI First and ROI Last evaluate sensitivity to evidence position, and Random Shuffle serves as a stronger structural stress test that disrupts inter-slice continuity. Evidence ablation separately evaluates diagnostic commitment under reduced annotated evidence.

To disentangle the two components of Label Swapping, we additionally evaluate instruction-only and schema-only variants on GPT-5.4 and Qwen3-VL-32B, representing proprietary and open-source generalist models. Complete results are provided in \cref{sec:appendix:textual_ablation} (\cref{tab:appendix-textual-disentanglement}).

\paragraph{Targeted Evidence Ablation.} 
To evaluate model behavior after removal of annotated lesion evidence, all slices intersecting expert consensus annotations of both the contrast-enhancing tumor core and peritumoral edema are systematically removed. The model is presented only with slices outside the expert-annotated tumor and edema regions. Consequently, no expert-annotated tumor or edema remains visible in the ablated input. With the primary annotated evidence removed, the expected response is \texttt{UNSURE} when the remaining evidence is insufficient.

To independently assess the sufficiency of the remaining visual evidence, a board-certified neuroradiologist performs a blinded audit of 30 ablated sequences sampled with balanced representation of diagnosis and modality. The reader selects \texttt{GBM}, \texttt{MET}, or \texttt{UNSURE} and separately indicates whether the remaining images are sufficient for a definitive GBM-versus-MET diagnosis. Sampling, blinding, and evaluation details are provided in \cref{sec:appendix:clinical_audit}. To test whether GPT-5.4's categorical outputs partly reflect reluctance to abstain, we repeat its baseline and slice-removal conditions with explicit guidance to select \texttt{UNSURE} when evidence is insufficient.

\begin{table}[t]
  \centering
  \small
  \setlength{\tabcolsep}{4pt}
  \renewcommand{\arraystretch}{1.05}
  \begin{tabularx}{\columnwidth}{ll >{\centering\arraybackslash}X >{\centering\arraybackslash}X}
    \toprule
    \textbf{Model} & \textbf{Mod.} &
    \textbf{Accuracy (\%)} &
    \textbf{Abstention (\%)} \\
    \midrule
    Gemini-2.5-Pro & T1CE & 80.0 &  0.0 \\
    Gemini-2.5-Pro & T2   & 63.3 & 10.0 \\
    GPT-5.4        & T1CE & 72.2 &  3.3 \\
    GPT-5.4        & T2   & 52.2 &  8.9 \\
    \midrule
    Qwen3-VL-32B   & T1CE & 60.0 &  3.3 \\
    Qwen3-VL-32B   & T2   & 45.6 & 28.9 \\
    MedGemma-27B   & T1CE & 63.6 &  2.3 \\
    MedGemma-27B   & T2   & 47.8 & 34.4 \\
    \midrule
    \multicolumn{4}{l}{\textit{Supervised Medical Imaging References}} \\
    Radiomics + SVM & T1CE & 71.0 & -- \\
    Radiomics + SVM & T2   & 79.0 & -- \\
    Radiomics + RF  & T1CE & 77.0 & -- \\
    Radiomics + RF  & T2   & 78.0 & -- \\
    \bottomrule
  \end{tabularx}
  \caption{Baseline clinical performance. Accuracy and abstention rates are measured on the unperturbed image sequences. Medical baselines serve as a reference.}
  \label{tab:baseline-performance}
\end{table}

\begin{table*}[t]
	\small
	\setlength{\tabcolsep}{4pt}          
	\renewcommand{\arraystretch}{1.05}   
	\begin{tabularx}{\textwidth}{ll *{6}{>{\centering\arraybackslash}X}}
		\toprule
		& & \multicolumn{4}{c}{\textbf{Visual Perturbations (Flip\%)}} & \multicolumn{2}{c}{\textbf{Textual Perturbations (Flip\%)}} \\
		\cmidrule(lr){3-6} \cmidrule(lr){7-8}
		\textbf{Model} & \textbf{Mod.} &
		\textbf{\begin{tabular}{@{}c@{}}Reverse\\Order\end{tabular}} &
		\textbf{\begin{tabular}{@{}c@{}}Random\\Shuffle\end{tabular}} &
		\textbf{\begin{tabular}{@{}c@{}}ROI\\First\end{tabular}} &
		\textbf{\begin{tabular}{@{}c@{}}ROI\\Last\end{tabular}} &
		\textbf{\begin{tabular}{@{}c@{}}Label\\Swap\end{tabular}} &
		\textbf{\begin{tabular}{@{}c@{}}Lexical\\Rephrase\end{tabular}} \\
		\midrule
		Gemini-2.5-Pro & T1CE & 11.1 &  \textbf{3.3} &  8.9 &  \textbf{4.4} &  6.7 &  \textbf{3.3} \\
		Gemini-2.5-Pro & T2   & 21.1 & 16.7 & 14.4 & 10.0 & 11.1 & 11.1 \\
		GPT-5.4        & T1CE &  \textbf{5.6} & 11.1 &  \textbf{3.3} &  7.8 &  \textbf{5.6} &  5.6 \\
		GPT-5.4        & T2   & 14.4 & 11.1 & 12.2 &  7.8 & 11.1 &  8.9 \\
		\midrule
		Qwen3-VL-32B   & T1CE &  \textbf{7.8} & \textbf{11.1} &  \textbf{10.0} & 10.0 & \textbf{14.4} &  8.9 \\
		Qwen3-VL-32B   & T2   & 16.7 & 27.8 & 11.1 &  \textbf{7.8} & 16.7 &  \textbf{5.6} \\
		MedGemma-27B   & T1CE & 23.9 & 19.3 & 21.6 & 13.6 & 59.1 & 19.3 \\
		MedGemma-27B   & T2   & 48.9 & 57.8 & 45.6 & 26.7 & 67.8 & 27.8 \\
		\bottomrule
	\end{tabularx}
	\caption{Diagnostic flip rates under controlled presentation perturbations. Values represent the percentage of subjects for which a model altered its baseline prediction under each perturbation. Bold indicates the most robust model configuration (lowest flip rate) across both modalities (T1CE and T2) within the proprietary (top four) and open-source (bottom four) model groups, respectively.}
	\label{tab:perturbation-flips}
\end{table*}

\subsection{Evaluation Metrics}
\label{sec:methodology:metrics}

To quantify diagnostic robustness, we use instance-level consistency metrics rather than relying solely on aggregate accuracy.

Our primary metric is \textbf{Flip Rate (\%)}, defined as the percentage of subjects for which a model's prediction differs from its own baseline output under a visual or textual perturbation. Baseline \texttt{UNSURE} cases are included, and any transition among \texttt{GBM}, \texttt{MET}, and \texttt{UNSURE} is counted as a flip.

Because Flip Rate may include incidental correction of a baseline error, we additionally report \textbf{Conditional Degradation Rate}, defined as the percentage of baseline-correct cases that become incorrect or \texttt{UNSURE} after perturbation.

For targeted evidence ablation, we report \textbf{Abstention Rate (\%)}, defined as the frequency of \texttt{UNSURE}, and its complement, the \textbf{Categorical Diagnosis Rate (\%)}, which measures diagnostic commitment after removal of the primary annotated lesion evidence.

\section{Results}
\label{sec:results}

\subsection{Baseline Clinical Performance}

We first establish the baseline clinical performance of the evaluated models on the standard, unperturbed anatomical sequences, as detailed in \cref{tab:baseline-performance}. While our core study critiques the over-reliance on static aggregate accuracy, reporting these initial figures establishes a default reference point of the models' capacities on the task.

Under baseline conditions, Gemini-2.5-Pro and GPT-5.4 achieve accuracies of 80.0\% and 72.2\% on the T1CE modality, respectively. The open-source Qwen3-VL-32B and domain-specific MedGemma-27B also demonstrate moderate baseline performance, particularly on the T1CE sequence, though their overall accuracies are generally lower than those of the proprietary generalist models.

However, these baseline figures may mask diagnostic instability. We therefore investigate whether predictions remain consistent under controlled changes in evidence presentation. For completeness, 95\% bootstrap confidence intervals are reported in the Appendix (\cref{tab:appendix-baseline-ablation-combined-ci,tab:appendix-flips-ci}).

To contextualize task difficulty, dedicated radiomics classifiers (SVM and RF) trained on expert ROI segmentations achieve baseline accuracies of 71.0\%--77.0\% on T1CE and 78.0\%--79.0\% on T2. These baselines confirm that the underlying structural MRI data contain discriminative radiological information for GBM-vs-MET differentiation. At the same time, this comparison highlights a key structural asymmetry: radiomics models rely on manual 3D ROI segmentation and supervised feature engineering, whereas VLMs operate zero-shot on raw, unsegmented slice sequences. The following analyses examine whether VLM baseline performance remains stable under changes in presentation ordering and linguistic framing.

\subsection{Sensitivity to Controlled Presentation Perturbations}

Under controlled presentation perturbations, prediction instability varies substantially across models and conditions (\cref{tab:perturbation-flips}). Sequence reversal induces diagnostic flips across all evaluated models, while random shuffling reveals additional sensitivity under a stronger aggregation stress test. Across the evaluated models, instability persists across three deterministic permutations, with mean, standard deviation, and worst-case flip rates reported in the Appendix (\cref{tab:appendix-multi-seed-shuffle}). We interpret this result as sensitivity to evidence aggregation rather than strict direction invariance, because random shuffling disrupts inter-slice continuity while preserving the available image content.

The proprietary models also exhibit sensitivity to these presentation variations. For instance, under simple sequence reversal, Gemini-2.5-Pro and GPT-5.4 alter their baseline predictions in 21.1\% and 14.4\% of T2 cases, respectively. This instability shows that model predictions can depend on slice presentation order, even when the available images remain unchanged.

Diagnostic instability also appears under textual framing changes. The label-swapping configuration reveals sensitivity to target-label ordering. This is most strikingly observed in the domain-specific MedGemma-27B model, which flipped its predictions in 59.1\% (T1CE) and 67.8\% (T2) of cases despite the visual evidence remaining entirely unchanged. While the generalist models demonstrate more resilience to label swapping compared to MedGemma-27B, they still exhibit measurable fluctuations (e.g., 14.4\% and 16.7\% for Qwen3-VL-32B).

Although some models exhibit seemingly minimal flip rates under certain configurations, such margins are far from negligible in safety-critical domains. For identical image content, a patient-level diagnosis should not change solely because sequence direction or target-label order is reversed. Such diagnostic volatility emphasizes the need for extensive verification protocols, as high prediction stability remains an important consideration for safe clinical integration.

A complementary analysis conditioned on baseline-correct cases confirms that instability is not explained only by incidental correction of initial errors. Conditional degradation reaches 55.8\% for MedGemma-27B under T2 Label Swapping, 24.4\% for Qwen3-VL-32B under T2 Random Shuffle, and 14.0\% for Gemini-2.5-Pro under T2 Reverse Order. Complete conditional-degradation results and absolute counts are provided in \cref{sec:appendix:conditional_degradation}, with categorical-versus-abstention flip breakdowns reported in \cref{sec:appendix:extended_results:flips}.

The disentangled textual ablation further shows that the joint intervention is not driven exclusively by one component. Instruction-only label reordering produces flip rates of up to 14.4\% for GPT-5.4 and 12.2\% for Qwen3-VL-32B, while schema-only reordering produces rates of up to 11.1\% and 8.9\%, respectively. Thus, both instructional wording and formal output ordering can independently affect predictions in the representative models evaluated. Full results are provided in the Appendix (\cref{tab:appendix-textual-disentanglement}).

Beyond the primary diagnostic outcome, presentation perturbations also affect secondary spatial predictions. Although sequence reversal changes only the order of the slices and does not mirror the individual images, GPT-5.4 exhibits left-to-right or right-to-left laterality changes in up to 20.5\% of T2 cases. MedGemma-27B also frequently loses spatial confidence under lexical rephrasing. Complete laterality-transition results and analysis are provided in \cref{sec:appendix:extended_results:laterality}.

\subsection{Flip Directionality}
\label{sec:results:flip_directionality} 

We further decompose diagnostic flips into degradation, incidental correction, and abstention, with the aggregated results reported in the Appendix (\cref{tab:flip-directionality}). This analysis distinguishes harmful perturbation effects from cases where an initially incorrect prediction changes by chance or where the model retreats to \texttt{UNSURE}. The results show that perturbation-induced instability is not merely a shift toward uncertainty and often includes categorical diagnostic changes. These proportions are computed conditionally on a flip occurring; full per-perturbation breakdowns are provided in \cref{sec:appendix:extended_results:flips}.

\subsection{Self-Reported Evidence vs. Diagnostic Robustness}
\label{sec:results:localization}

In addition to the primary diagnosis, models reported the indices of slices containing task-critical lesions. Their self-reported localization IoU remains stable across perturbations despite the observed diagnostic flips; results are reported in the Appendix (\cref{tab:localization-iou}). However, autoregressive reporting of evidence locations does not necessarily reflect the causal mechanism underlying the final classification. The coexistence of stable reported localization and unstable diagnosis is therefore consistent with a behavioral dissociation between reported evidence localization and diagnostic output.

\begin{table}[t]
  \centering
  \small
  \setlength{\tabcolsep}{4pt}
  \renewcommand{\arraystretch}{1.05}
  \begin{tabularx}{\columnwidth}{ll >{\centering\arraybackslash}X >{\centering\arraybackslash}X}
    \toprule
    \textbf{Model} & \textbf{Mod.} &
    \textbf{Abstention (\%)} &
    \textbf{Categorical (\%)} \\
    \midrule
    Gemini-2.5-Pro & T1CE & 41.1 & 58.9 \\
    Gemini-2.5-Pro & T2   & 77.8 & 22.2 \\
    GPT-5.4        & T1CE & 73.3 & 26.7 \\
    GPT-5.4        & T2   & \textbf{90.0} & \textbf{10.0} \\
    \midrule
    Qwen3-VL-32B   & T1CE & 68.9 & 31.1 \\
    Qwen3-VL-32B   & T2   & \textbf{98.9} &  \textbf{1.1} \\
    MedGemma-27B   & T1CE & 23.9 & 76.1 \\
    MedGemma-27B   & T2   & 72.2 & 27.8 \\
    \bottomrule
  \end{tabularx}
    \caption{Evidence-ablation evaluation. Categorical denotes predictions issued after removal of annotated lesion slices.}
  \label{tab:negative-control}
\end{table}

\subsection{Diagnostic Overcommitment Under Evidence Ablation}

Our evidence-ablation evaluation highlights a notable diagnostic overcommitment pattern (\cref{tab:negative-control}). Our expert annotations deliberately encompass both the enhancing tumor core and the peritumoral edema. Once these slices are removed, the input no longer contains the expert-annotated tumor and edema slices. When the residual images are insufficient for a definitive GBM-versus-MET diagnosis, the expected behavior is abstention (\texttt{UNSURE}); the blinded audit supports this assessment for the audited subset. We refer to categorical prediction under these conditions as diagnostic overcommitment, reflecting commitment despite removal of the primary annotated evidence and the availability of an abstention option.

Instead, the results reveal a strong propensity among certain models to issue categorical diagnoses. Gemini-2.5-Pro and MedGemma-27B produce categorical outputs in 58.9\% and 76.1\% of T1CE cases, respectively, despite the removal of expert-annotated lesion slices. This behavior highlights a safety concern, as models can generate categorical diagnoses after the primary expert-annotated tumor and edema evidence has been removed.

The blinded specialist audit independently supports the expected abstention behavior. The neuroradiologist selected \texttt{UNSURE} for all 30 audited ablated sequences and judged the remaining visual evidence insufficient for a definitive GBM-versus-MET diagnosis in every case. For the audited subset, this clinical assessment supports interpreting categorical model outputs under evidence ablation as diagnostic overcommitment.

It is worth noting, however, that abstention behaviors vary significantly across modalities and model architectures. GPT-5.4 and Qwen3-VL-32B demonstrate a more conservative abstention pattern on the T2 modality, outputting \texttt{UNSURE} in 90.0\% and 98.9\% of cases, respectively.

Explicit abstention guidance leaves GPT-5.4 baseline behavior nearly unchanged but increases abstention after slice removal from 73.3\% to 80.0\% on T1CE and from 90.0\% to 97.8\% on T2. Thus, diagnostic commitment is partly instruction-sensitive, although 20.0\% of the guided T1CE cases remain categorical; full results are provided in \cref{sec:appendix:safety_calibration}.

\subsection{Cross-Task External Evaluation}
\label{sec:results:external}

A similar presentation-sensitivity pattern appears on the external BraTS cohort under a distinct HGG-versus-LGG task. Under sequence reversal, GPT-5.4 changes 8.0\% of T1CE predictions and 14.0\% of T2 predictions, while Qwen3-VL-32B changes 14.0\% and 26.0\%, respectively. Target-label reordering produces flip rates of 3.0\%--12.0\% for GPT-5.4 and 6.0\%--15.0\% for Qwen3-VL-32B. These results extend the central presentation-sensitivity pattern to a distinct dataset and diagnostic task. Full baseline and perturbation results are reported in the Appendix (\cref{tab:appendix-external-validation}).

\section{Conclusion} 
\label{sec:conclusion} 

We evaluated four generalist and medically adapted VLMs on a histopathology-confirmed brain MRI cohort. Despite competitive baseline performance, controlled changes in slice order and textual framing altered diagnoses and degraded initially correct predictions. Some models also produced categorical outputs after the annotated lesion evidence was removed, despite specialist assessment that the remaining images were insufficient for diagnosis. A similar presentation-sensitivity pattern was also observed on an external cohort under a distinct glioma-grading task. Overall, prediction stability and evidence-sensitive abstention should complement canonical accuracy before VLMs are considered for safety-critical multimodal workflows.

\section*{Limitations}
\label{sec:limitations}

First, evaluating complete MRI volumes remains challenging for some open-source VLMs due to multimodal context-length limitations. Two exceptionally long sequences exceeded the operational limits of MedGemma-27B and were therefore excluded from its evaluation subset. To maintain a controlled comparison, we avoided model-specific interventions such as truncation, subsampling, or image compositing.

Second, our evaluation is necessarily black-box. Because leading proprietary VLMs do not expose internal representations or comparable reasoning traces, observed instabilities cannot be directly attributed to specific internal mechanisms. We therefore focus on diagnostic behavior and prediction consistency as observable model outputs.

Third, the perturbations considered in this study intentionally isolate specific presentation factors, including slice ordering, evidence position, and textual framing. This controlled design enables precise measurement of prediction stability, although real-world clinical imaging environments may involve multiple interacting sources of variability.

\section*{Ethical Considerations}
\label{sec:ethical}

The retrospective use of clinical data for this research was approved by the relevant Institutional Review Board and Ethics Committee. Because this investigation is based exclusively on historical clinical data, we did not collect prospective patient data, and no direct clinical interactions occurred. Before conducting any computational evaluations, we verified a strict de-identification protocol by stripping all DICOM metadata, headers, and protected health information, preserving only the derived image representations. To confirm that no personal identifiers remained, we performed manual audits of the files, ensuring this retrospective study introduces no new risks to participants in accordance with applicable data-protection standards.

The VLMs evaluated in this work are intended strictly for academic research and are not designed to offer clinical recommendations, diagnoses, or active medical advice. By demonstrating prediction sensitivity and fragility in current VLMs under benign presentation changes, this study serves to caution against over-reliance on AI within safety-critical clinical settings. Exposing these vulnerabilities helps mitigate potential dual-use concerns and underscores that the evaluated configurations should not be used autonomously in this diagnostic setting.

To support scientific reproducibility and open-science initiatives, our evaluation data and code will be made publicly available in de-identified form upon acceptance. These resources are accompanied by documentation outlining their scientific scope, intended research applications, and known limitations. Our evaluation data are not designed for clinical decision-making, commercial deployment, or diagnostic use, and are governed by terms of use structured to ensure responsible scientific reuse and prevent potential misuse.

\bibliography{custom}

\clearpage
\appendix
\crefalias{section}{appendix}
\crefalias{subsection}{appendix}

\section{Extended Dataset and Clinical Details}
\label{sec:appendix:dataset}

This work utilizes a retrospectively collected, multi-center brain MRI dataset. This clinical cohort provides a structured and rigorously annotated environment that inherently supports the controlled manipulation of visual evidence without compromising medical validity. The dataset comprises axial slice sequences obtained during routine clinical workflows. Every subject in the cohort has a histopathologically confirmed definitive diagnosis, and the imaging data is paired with expert-delineated spatial annotations.

\subsection{Clinical Definitions}
\label{sec:appendix:clinical_definitions}

To contextualize the neuro-oncological task for the NLP community, we outline the primary clinical concepts pertinent to our evaluation:

\paragraph{Magnetic Resonance Imaging (MRI).}
MRI is the primary non-invasive imaging modality for diagnosing and characterizing intra-axial brain tumors. It provides exceptional soft-tissue contrast. Because clinical decision-making typically relies on standard structural sequences rather than advanced functional imaging, our dataset focuses on the two most critical routinely acquired modalities.

\paragraph{Diagnostic Lesion.}
In the context of brain tumors, a lesion encompasses the macroscopic pathological tissue alterations, including the primary tumor mass, necrotic cores, and surrounding peritumoral edema. In our perturbation framework, slices intersecting the expert-annotated tumor or edema regions constitute the task-critical visual evidence.

\paragraph{T1-weighted contrast-enhanced (T1CE).}
This sequence is acquired following the intravenous administration of a gadolinium-based contrast agent. It is highly sensitive to the breakdown of the blood-brain barrier, allowing clinicians (and models) to visualize the actively enhancing components and internal architecture of the tumor.

\paragraph{T2-weighted (T2).}
T2 sequences are highly sensitive to tissue water content. In neuro-oncology, they are indispensable for identifying non-enhancing pathological regions, particularly the hyperintense peritumoral edema that extends beyond the visible boundaries of the contrast-enhancing core.

\paragraph{Glioblastoma (\texttt{GBM}) vs. Brain Metastasis (\texttt{MET}).}
Glioblastoma is the most prevalent and aggressive primary malignant brain tumor in adults. Conversely, brain metastases are secondary tumors that have spread to the central nervous system from systemic primary cancers (e.g., lung or breast). Differentiating between the two---especially when they present as solitary, contrast-enhancing masses---is a notoriously difficult radiological task, making it an ideal testbed for evaluating rigorous multimodal reasoning.

\begin{table}[t]
	\centering
	\small
	\setlength{\tabcolsep}{4pt}
	\renewcommand{\arraystretch}{1.05}
	\begin{tabular*}{\columnwidth}{@{\extracolsep{\fill}} lcc @{}}
		\toprule
		\textbf{Metric} &
		\textbf{\begin{tabular}{@{}c@{}}Median\\(IQR)\end{tabular}} &
		\textbf{\begin{tabular}{@{}c@{}}Range\\(min--max)\end{tabular}} \\
		\midrule
		T1CE slices / subject &
		20 (19--24) & 15--160 \\
		T2 slices / subject &
		20 (18--22) & 15--62 \\
		\midrule
		Lesion slices / subject (T1CE) &
		8 (6--11) & 3--34 \\
		Lesion slices / subject (T2) &
		11 (8.25--13) & 4--23 \\
		\midrule
		Lesion slice fraction (T1CE) &
		0.39 (0.27--0.53) & 0.15--0.85 \\
		Lesion slice fraction (T2) &
		0.55 (0.40--0.67) & 0.20--0.89 \\
		\bottomrule
	\end{tabular*}
	\caption{Summary statistics for the MRI slice sequences. Lesion slices denote the images identified by expert neuroradiologists as containing task-critical pathological evidence.}
	\label{tab:appendix-dataset-summary}
\end{table}

\subsection{Annotation and Ground Truth Formulation}
\label{sec:appendix:annotation}

To establish the spatial coordinates necessary for our perturbation engine, lesion delineation was conducted through a rigorous dual-reader protocol. Two board-certified neuroradiologists independently segmented the tumor and associated edema at the pixel level across all axial slices. Disagreements were subsequently resolved in a joint consensus session to produce a single definitive binary mask per sequence. In our pipeline, these masks serve solely to identify which sequence slices contain task-critical evidence and which do not.

It is imperative to note that the subject-level diagnostic labels (\texttt{GBM} vs. \texttt{MET}) are entirely independent of these radiological masks. Ground-truth labels were established strictly via histopathological examination following surgical resection or biopsy, eliminating any subjective radiological bias from the evaluation targets.

\begin{table*}[t!]
  \centering
  
  \small
  \setlength{\tabcolsep}{4pt}
  \renewcommand{\arraystretch}{1.05}
  \begin{tabularx}{\textwidth}{ll c *{4}{>{\centering\arraybackslash}X}}
    \toprule
    & & \multicolumn{3}{c}{\textbf{Baseline Performance (95\% CI)}} & \multicolumn{2}{c}{\textbf{Evidence-ablation Evaluation (95\% CI)}} \\
    \cmidrule(lr){4-5} \cmidrule(lr){6-7}
    \textbf{Model} & \textbf{Mod.} & \textbf{N} &
    \textbf{Accuracy} &
    \textbf{Abstention} &
    \textbf{Abstention} &
    \textbf{Categorical} \\
    \midrule
    Gemini-2.5-Pro & T1CE & 90 & 80.0 [72.2, 88.9] &  0.0 [0.0, 0.0]   & 41.1 [31.1, 51.1] & 58.9 [48.9, 68.9] \\
    Gemini-2.5-Pro & T2   & 90 & 63.3 [52.2, 74.4] & 10.0 [4.4, 16.7]  & 77.8 [68.9, 86.7] & 22.2 [14.4, 31.1] \\
    GPT-5.4        & T1CE & 90 & 72.2 [62.2, 81.1] &  3.3 [0.0, 7.8]   & 73.3 [63.3, 82.2] & 26.7 [17.8, 36.7] \\
    GPT-5.4        & T2   & 90 & 52.2 [42.2, 62.2] &  8.9 [3.3, 15.6]  & 90.0 [83.3, 95.6] & 10.0 [4.4, 16.7] \\
    \midrule
    Qwen3-VL-32B   & T1CE & 90 & 60.0 [50.0, 70.0] &  3.3 [0.0, 7.8]   & 68.9 [58.9, 78.9] & 31.1 [22.2, 41.1] \\
    Qwen3-VL-32B   & T2   & 90 & 45.6 [35.6, 55.6] & 28.9 [20.0, 38.9] & 98.9 [96.7, 100.0] & 1.1 [0.0, 3.3] \\
    MedGemma-27B   & T1CE & 88 & 63.6 [53.4, 73.9] &  2.3 [0.0, 5.7]   & 23.9 [15.9, 31.8] & 76.1 [67.0, 85.2] \\
    MedGemma-27B   & T2   & 90 & 47.8 [37.8, 57.8] & 34.4 [25.6, 44.4] & 72.2 [62.2, 81.1] & 27.8 [17.8, 36.7] \\
    \bottomrule
  \end{tabularx}
  \caption{Baseline performance and evidence-ablation evaluation, presented with 95\% bootstrap confidence intervals.}
  \label{tab:appendix-baseline-ablation-combined-ci}

  \vspace{1cm}

\scriptsize
  \setlength{\tabcolsep}{3pt}          
  \renewcommand{\arraystretch}{1.1}   
  \begin{tabularx}{\textwidth}{ll *{6}{>{\centering\arraybackslash}X}}
    \toprule
    & & \multicolumn{4}{c}{\textbf{Visual Perturbations (Flip\% [95\% CI])}} & \multicolumn{2}{c}{\textbf{Textual Perturbations (Flip\% [95\% CI])}} \\
    \cmidrule(lr){3-6} \cmidrule(lr){7-8}
    \textbf{Model} & \textbf{Mod.} &
    \textbf{Reverse Order} &
    \textbf{Random Shuffle} &
    \textbf{ROI First} &
    \textbf{ROI Last} &
    \textbf{Label Swap} &
    \textbf{Lexical Rephrase} \\
    \midrule
    Gemini-2.5-Pro & T1CE & 11.1 [4.4, 17.8] &  3.3 [0.0, 7.8]  &  8.9 [3.3, 14.4] &  4.4 [1.1, 8.9]  &  6.7 [2.2, 12.2] &  3.3 [0.0, 6.7] \\
    Gemini-2.5-Pro & T2   & 21.1 [13.3, 30.0]& 16.7 [10.0, 24.4]& 14.4 [7.8, 22.2] & 10.0 [4.4, 16.7] & 11.1 [4.4, 17.8] & 11.1 [5.6, 17.8] \\
    GPT-5.4        & T1CE &  5.6 [1.1, 11.1] & 11.1 [5.6, 17.8] &  3.3 [0.0, 7.8]  &  7.8 [3.3, 14.4] &  5.6 [1.1, 11.1] &  5.6 [1.1, 11.1] \\
    GPT-5.4        & T2   & 14.4 [7.8, 22.2] & 11.1 [5.6, 18.9] & 12.2 [5.6, 20.0] &  7.8 [2.2, 14.4] & 11.1 [5.6, 17.8] &  8.9 [3.3, 15.6] \\
    \midrule
    Qwen3-VL-32B   & T1CE &  7.8 [2.2, 14.4] & 11.1 [5.6, 17.8] & 10.0 [4.4, 16.7] & 10.0 [4.4, 16.7] & 14.4 [7.8, 22.2] &  8.9 [3.3, 14.4] \\
    Qwen3-VL-32B   & T2   & 16.7 [10.0, 24.4]& 27.8 [20.0, 36.7]& 11.1 [4.4, 17.8] &  7.8 [2.2, 13.3] & 16.7 [8.9, 24.5] &  5.6 [1.1, 11.1] \\
    MedGemma-27B   & T1CE & 23.9 [14.8, 33.0]& 19.3 [11.4, 28.4]& 21.6 [13.6, 30.7]& 13.6 [6.8, 21.6] & 59.1 [48.9, 69.3]& 19.3 [11.4, 28.4] \\
    MedGemma-27B   & T2   & 48.9 [37.8, 58.9]& 57.8 [47.8, 68.9]& 45.6 [35.6, 55.6]& 26.7 [18.9, 36.7]& 67.8 [58.9, 77.8]& 27.8 [18.9, 36.7] \\
    \bottomrule
  \end{tabularx}
  \caption{Diagnostic flip rates under perturbations, presented with 95\% bootstrap confidence intervals.}
  \label{tab:appendix-flips-ci}

  \vspace{1cm}

\scriptsize
  \setlength{\tabcolsep}{4pt}          
  \renewcommand{\arraystretch}{1.1}   
  \begin{tabularx}{\textwidth}{ll *{6}{>{\centering\arraybackslash}X}}
    \toprule
    \textbf{Model} & \textbf{Mod.} & \textbf{Reverse Order} & \textbf{Random Shuffle} & \textbf{ROI First} & \textbf{ROI Last} & \textbf{Label Swap} & \textbf{Lexical Rephrase} \\
    \midrule
    Gemini-2.5-Pro & T1CE & 60/40/0  & 67/33/0  & 75/25/0  & 50/50/0  & 50/50/0  & 100/0/0  \\
    Gemini-2.5-Pro & T2   & 47/32/21 & 27/47/27 & 15/62/23 & 22/56/22 & 30/30/40 & 40/40/20 \\
    GPT-5.4        & T1CE & 40/60/0  & 60/40/0  & 67/33/0  & 29/71/0  & 40/60/0  & 20/80/0  \\
    GPT-5.4        & T2   & 23/77/0  & 10/90/0  & 18/82/0  & 14/86/0  & 20/80/0  & 25/75/0  \\
    \midrule
    Qwen3-VL-32B   & T1CE & 57/29/14 & 40/50/10 & 33/67/0  & 44/33/22 & 54/23/23 & 25/50/25 \\
    Qwen3-VL-32B   & T2   & 33/47/20 & 28/44/28 & 40/60/0  & 14/86/0  &  7/27/67 & 40/60/0  \\
    MedGemma-27B   & T1CE & 48/19/33 & 47/53/0  & 53/42/5  & 17/50/33 & 40/42/17 & 24/35/41 \\
    MedGemma-27B   & T2   & 45/34/20 & 46/46/8  & 54/44/2  & 17/17/67 & 57/31/11 & 16/24/60 \\
    \bottomrule
  \end{tabularx}
  \caption{Detailed flip breakdown per perturbation. Values denote the percentage of flips that resulted in (Correct$\rightarrow$Wrong / Wrong$\rightarrow$Correct / To UNSURE) respectively.}
  \label{tab:appendix-detailed-flips}

  \vspace{1cm}

  \small
  \setlength{\tabcolsep}{4pt}
  \renewcommand{\arraystretch}{1.05}
  \begin{tabularx}{\textwidth}{ll *{7}{>{\centering\arraybackslash}X}}
    \toprule
    & & & \multicolumn{4}{c}{\textbf{Visual Perturbations}} & \multicolumn{2}{c}{\textbf{Textual Perturbations}} \\
    \cmidrule(lr){4-7} \cmidrule(lr){8-9}
    \textbf{Model} & \textbf{Mod.} & \textbf{Baseline} &
    \textbf{Reverse Order} & \textbf{Random Shuffle} & \textbf{ROI First} & \textbf{ROI Last} & \textbf{Label Swap} & \textbf{Lexical Rephrase} \\
    \midrule
    Gemini-2.5-Pro & T1CE & 0.769 & 0.774 & 0.762 & 0.803 & 0.789 & 0.762 & 0.758 \\
    Gemini-2.5-Pro & T2   & 0.801 & 0.781 & 0.791 & 0.829 & 0.805 & 0.803 & 0.785 \\
    GPT-5.4        & T1CE & 0.719 & 0.758 & 0.733 & 0.707 & 0.783 & 0.718 & 0.711 \\
    GPT-5.4        & T2   & 0.772 & 0.792 & 0.770 & 0.772 & 0.799 & 0.781 & 0.773 \\
    \midrule
    Qwen3-VL-32B   & T1CE & 0.683 & 0.689 & 0.666 & 0.707 & 0.691 & 0.675 & 0.661 \\
    Qwen3-VL-32B   & T2   & 0.630 & 0.606 & 0.604 & 0.646 & 0.626 & 0.617 & 0.646 \\
    MedGemma-27B   & T1CE & 0.430 & 0.409 & 0.423 & 0.414 & 0.458 & 0.433 & 0.443 \\
    MedGemma-27B   & T2   & 0.565 & 0.523 & 0.554 & 0.533 & 0.597 & 0.567 & 0.580 \\
    \bottomrule
  \end{tabularx}
  \caption{Spatial localization performance. Values denote the average Intersection over Union (IoU) between the model-reported evidence slices and the ground-truth lesion annotations across all test configurations.}
  \label{tab:localization-iou}

\end{table*}

\begin{table*}[t]
	\centering
	\small
	\setlength{\tabcolsep}{2pt}          
	\renewcommand{\arraystretch}{1.3}   
	\begin{tabularx}{\textwidth}{ll *{6}{>{\centering\arraybackslash}X}}
		\toprule
		& & \multicolumn{4}{c}{\textbf{Visual Perturbations (\%)}} & \multicolumn{2}{c}{\textbf{Textual Perturbations (\%)}} \\
		\cmidrule(lr){3-6} \cmidrule(lr){7-8}
		\textbf{Model} & \textbf{Mod.} &
		\textbf{Reverse Order} &
		\textbf{Random Shuffle} &
		\textbf{ROI First} &
		\textbf{ROI Last} &
		\textbf{Label Swap} &
		\textbf{Lexical Rephrase} \\
		\midrule
		Gemini-2.5-Pro & T1CE & 10.0 \newline \scriptsize [2.2/0.0/4.4] & 10.0 \newline \scriptsize [2.2/0.0/2.2] & 6.7 \newline \scriptsize [2.2/0.0/1.1] & 6.7 \newline \scriptsize [3.3/0.0/1.1] & 2.2 \newline \scriptsize [2.2/0.0/0.0] & 8.9 \newline \scriptsize [2.2/0.0/3.3] \\
		Gemini-2.5-Pro & T2   & 13.5 \newline \scriptsize [0.0/1.1/7.9] & 10.1 \newline \scriptsize [0.0/1.1/6.7] & 9.0 \newline \scriptsize [0.0/0.0/6.7] & 9.0 \newline \scriptsize [0.0/1.1/4.5] & 5.6 \newline \scriptsize [0.0/0.0/2.2] & 12.4 \newline \scriptsize [0.0/0.0/6.7] \\
		\midrule
		GPT-5.4        & T1CE & 22.1 \newline \scriptsize [15.1/2.3/2.3] & 14.0 \newline \scriptsize [9.3/1.2/2.3] & 9.3 \newline \scriptsize [8.1/0.0/0.0] & 16.3 \newline \scriptsize [10.5/1.2/4.7] & 11.6 \newline \scriptsize [9.3/0.0/1.2] & 14.0 \newline \scriptsize [11.6/0.0/2.3] \\
		GPT-5.4        & T2   & 28.4 \newline \scriptsize [20.5/2.3/4.5] & 19.3 \newline \scriptsize [11.4/2.3/4.5] & 20.5 \newline \scriptsize [13.6/1.1/4.5] & 15.9 \newline \scriptsize [8.0/2.3/5.7] & 13.6 \newline \scriptsize [9.1/1.1/3.4] & 12.5 \newline \scriptsize [9.1/0.0/3.4] \\
		\midrule
		Qwen3-VL-32B   & T1CE & 18.4 \newline \scriptsize [9.2/2.3/4.6] & 21.6 \newline \scriptsize [11.4/2.3/5.7] & 10.2 \newline \scriptsize [5.7/1.1/1.1] & 11.4 \newline \scriptsize [5.7/3.4/1.1] & 9.1 \newline \scriptsize [5.7/1.1/2.3] & 14.8 \newline \scriptsize [8.0/2.3/4.5] \\
		Qwen3-VL-32B   & T2   & 15.0 \newline \scriptsize [6.2/5.0/2.5] & 10.0 \newline \scriptsize [5.0/1.2/2.5] & 3.8 \newline \scriptsize [1.2/1.2/0.0] & 8.8 \newline \scriptsize [1.2/5.0/2.5] & 8.8 \newline \scriptsize [3.8/3.8/1.2] & 10.0 \newline \scriptsize [2.5/6.2/1.2] \\
		\midrule
		MedGemma-27B   & T1CE & 64.3 \newline \scriptsize [3.6/60.7/0.0] & 46.4 \newline \scriptsize [28.6/17.9/0.0] & 35.7 \newline \scriptsize [17.9/17.9/0.0] & 21.4 \newline \scriptsize [17.9/3.6/0.0] & 35.7 \newline \scriptsize [7.1/28.6/0.0] & 57.1 \newline \scriptsize [7.1/50.0/0.0] \\
		MedGemma-27B   & T2   & 65.9 \newline \scriptsize [11.4/54.5/0.0] & 43.2 \newline \scriptsize [18.2/25.0/0.0] & 36.4 \newline \scriptsize [22.7/13.6/0.0] & 25.0 \newline \scriptsize [18.2/6.8/0.0] & 34.1 \newline \scriptsize [22.7/11.4/0.0] & 52.3 \newline \scriptsize [6.8/45.5/0.0] \\
		\bottomrule
	\end{tabularx}
    \caption{Changes in model predictions for lesion location (Left, Right, or Bilateral) under perturbations. Values are presented as \textbf{Total Change \% [Left/Right Swap / Lost Confidence / Shift to Bilateral]}, with bracketed values reporting selected transition types.}
	\label{tab:appendix-laterality}
\end{table*}

\begin{table*}[t]
  \centering
  \small
  \setlength{\tabcolsep}{4pt}
  \renewcommand{\arraystretch}{1.05}
  \begin{tabularx}{\textwidth}{ll *{5}{>{\centering\arraybackslash}X}}
    \toprule
    \textbf{Model} & \textbf{Mod.} & \textbf{Seed 42 (\%)} & \textbf{Seed 1998 (\%)} & \textbf{Seed 2026 (\%)} & \textbf{Mean $\pm$ SD (\%)} & \textbf{Worst-Case (\%)} \\
    \midrule
    Gemini-2.5-Pro & T1CE &  3.3 & 14.4 &  8.8 &  8.8 $\pm$ 5.6 & 14.4 \\
    Gemini-2.5-Pro & T2   & 16.7 & 22.2 & 16.7 & 18.5 $\pm$ 3.2 & 22.2 \\
    \midrule
    GPT-5.4        & T1CE & 11.1 &  8.9 &  5.6 &  8.5 $\pm$ 2.8 & 11.1 \\
    GPT-5.4        & T2   & 11.1 & 11.1 & 14.4 & 12.2 $\pm$ 1.9 & 14.4 \\
    \midrule
    Qwen3-VL-32B   & T1CE & 11.1 & 13.3 & 15.6 & 13.3 $\pm$ 2.2 & 15.6 \\
    Qwen3-VL-32B   & T2   & 27.8 & 21.1 & 17.8 & 22.2 $\pm$ 5.1 & 27.8 \\
    \midrule
    MedGemma-27B   & T1CE & 19.3 & 21.6 & 19.3 & 20.1 $\pm$ 1.3 & 21.6 \\
    MedGemma-27B   & T2   & 57.8 & 51.1 & 48.9 & 52.6 $\pm$ 4.6 & 57.8 \\
    \bottomrule
  \end{tabularx}
  \caption{Multi-seed evaluation of the Random Shuffle perturbation across three independent deterministic seeds (42, 1998, 2026), reporting Mean $\pm$ Standard Deviation and Worst-Case (Maximum) flip rates.}
  \label{tab:appendix-multi-seed-shuffle}
\end{table*}

\section{Extended Quantitative Results}
\label{sec:appendix:extended_results}

This section provides the supplementary quantitative metrics and granular breakdowns that support the findings presented in the main text. 

\subsection{Confidence Intervals for Diagnostic Metrics}
\label{sec:appendix:extended_results:cis}

To establish the statistical reliability of our evaluations, we compute 95\% bootstrap confidence intervals (CIs) for the primary metrics. \cref{tab:appendix-baseline-ablation-combined-ci} reports the CIs for the baseline clinical performance and the targeted evidence-ablation experiments. Correspondingly, \cref{tab:appendix-flips-ci} provides the CIs for the diagnostic flip rates under both visual and textual perturbations.

\subsection{Multi-Seed Shuffling Evaluation}
\label{sec:appendix:extended_results:multiseed}

To determine whether shuffle sensitivity is specific to a single permutation, we evaluate Random Shuffle using three deterministic seeds (42, 1998, and 2026). As shown in \cref{tab:appendix-multi-seed-shuffle}, instability persists across permutations. MedGemma-27B on T2 exhibits a mean flip rate of $52.6\% \pm 4.6\%$ and a worst-case rate of $57.8\%$, while Qwen3-VL-32B on T2 exhibits a mean rate of $22.2\% \pm 5.1\%$. Thus, the observed instability is not specific to the original permutation.

\subsection{Conditional Degradation of Baseline-Correct Diagnoses}
\label{sec:appendix:conditional_degradation}

Because overall Flip Rate includes harmful degradation, incidental correction, and transitions involving \texttt{UNSURE}, we additionally condition on subjects diagnosed correctly at baseline. Conditional Degradation Rate measures the proportion of these cases that become incorrect or \texttt{UNSURE} after perturbation.

\begin{table*}[t]
    \centering
    \scriptsize
    \setlength{\tabcolsep}{3pt}
    \renewcommand{\arraystretch}{1.12}
    \begin{tabularx}{\textwidth}{ll *{6}{>{\centering\arraybackslash}X}}
        \toprule
        \textbf{Model} & \textbf{Mod.} &
        \textbf{Reverse} &
        \textbf{Shuffle} &
        \textbf{ROI First} &
        \textbf{ROI Last} &
        \textbf{Label Swap} &
        \textbf{Rephrase} \\
        \midrule
        Gemini-2.5-Pro & T1CE & 6/72 (8.3) & 2/72 (2.8) & 6/72 (8.3) & 2/72 (2.8) & 3/72 (4.2) & 3/72 (4.2) \\
        Gemini-2.5-Pro & T2   & 8/57 (14.0) & 4/57 (7.0) & 4/57 (7.0) & 3/57 (5.3) & 5/57 (8.8) & 3/57 (5.3) \\
        GPT-5.4 & T1CE & 2/65 (3.1) & 6/65 (9.2) & 2/65 (3.1) & 2/65 (3.1) & 1/65 (1.5) & 1/65 (1.5) \\
        GPT-5.4 & T2 & 3/47 (6.4) & 1/47 (2.1) & 2/47 (4.3) & 1/47 (2.1) & 2/47 (4.3) & 2/47 (4.3) \\
        \midrule
        Qwen3-VL-32B & T1CE & 5/54 (9.3) & 5/54 (9.3) & 3/54 (5.6) & 4/54 (7.4) & 10/54 (18.5) & 3/54 (5.6) \\
        Qwen3-VL-32B & T2 & 4/41 (9.8) & 10/41 (24.4) & 2/41 (4.9) & 1/41 (2.4) & 5/41 (12.2) & 1/41 (2.4) \\
        MedGemma-27B & T1CE & 15/56 (26.8) & 8/56 (14.3) & 11/56 (19.6) & 3/56 (5.4) & 25/56 (44.6) & 7/56 (12.5) \\
        MedGemma-27B & T2 & 17/43 (39.5) & 18/43 (41.9) & 12/43 (27.9) & 15/43 (34.9) & 24/43 (55.8) & 15/43 (34.9) \\
        \bottomrule
    \end{tabularx}
    \caption{Conditional Degradation Rate under all perturbations. Cells report degraded cases/baseline-correct cases, followed by percentage. Degradation includes transitions from a correct baseline diagnosis to either an incorrect categorical diagnosis or \texttt{UNSURE}.}
    \label{tab:appendix-conditional-degradation}
\end{table*}

\subsection{Target-Label Reordering Ablation} \label{sec:appendix:textual_ablation}

To separate the effects of target-label order in the diagnostic instruction and output schema, we evaluate joint, instruction-only, and schema-only reordering on GPT-5.4 and Qwen3-VL-32B. Visual inputs remain identical across all conditions. All predictions are evaluated using the canonical class identities \texttt{GBM}, \texttt{MET}, and \texttt{UNSURE}; reordering label presentation does not alter the ground-truth mapping or output normalization.

\begin{table}[t]

    \centering
    \small
    \setlength{\tabcolsep}{2pt}
    \renewcommand{\arraystretch}{1.08}
    \begin{tabularx}{\columnwidth}{ll *{3}{>{\centering\arraybackslash}X}}
        \toprule
        \textbf{Model} & \textbf{Mod.} &
        \textbf{Joint} &
        \textbf{Instruction} &
        \textbf{Schema} \\
        \midrule
        GPT-5.4 & T1CE & 5.6 & 7.8 & 4.4 \\
        GPT-5.4 & T2   & 11.1 & 14.4 & 11.1 \\
        Qwen3-VL-32B & T1CE & 14.4 & 12.2 & 8.9 \\
        Qwen3-VL-32B & T2   & 16.7 & 10.0 & 8.9 \\
        \bottomrule
    \end{tabularx}
    \caption{Flip rates under joint, instruction-only, and schema-only target-label reordering. Visual inputs remain identical across all conditions.}
    \label{tab:appendix-textual-disentanglement}
\end{table}

\begin{table*}[t]
    \centering
    \small
    \setlength{\tabcolsep}{5pt}
    \renewcommand{\arraystretch}{1.05}
    \begin{tabularx}{\textwidth}{ll *{6}{>{\centering\arraybackslash}X}}
        \toprule
        & & \multicolumn{3}{c}{\textbf{Visual Flip Breakdown (\%)}} & \multicolumn{3}{c}{\textbf{Textual Flip Breakdown (\%)}} \\
        \cmidrule(lr){3-5} \cmidrule(lr){6-8}
        \textbf{Model} & \textbf{Mod.} &
        \makecell{\textbf{Correct}\\\textbf{$\rightarrow$ Wrong}} & 
        \makecell{\textbf{Wrong}\\\textbf{$\rightarrow$ Correct}} & 
        \makecell{\textbf{To}\\\textbf{UNSURE}} &
        \makecell{\textbf{Correct}\\\textbf{$\rightarrow$ Wrong}} & 
        \makecell{\textbf{Wrong}\\\textbf{$\rightarrow$ Correct}} & 
        \makecell{\textbf{To}\\\textbf{UNSURE}} \\
        \midrule
        Gemini-2.5-Pro & T1CE & 64.0 & 36.0 &  0.0 & 66.7 & 33.3 &  0.0 \\
        Gemini-2.5-Pro & T2   & 30.4 & 46.4 & 23.2 & 35.0 & 35.0 & 30.0 \\
        GPT-5.4        & T1CE & 48.0 & 52.0 &  0.0 & 30.0 & 70.0 &  0.0 \\
        GPT-5.4        & T2   & 17.1 & 82.9 &  0.0 & 22.2 & 77.8 &  0.0 \\
        \midrule
        Qwen3-VL-32B   & T1CE & 42.9 & 45.7 & 11.4 & 42.9 & 33.3 & 23.8 \\
        Qwen3-VL-32B   & T2   & 29.8 & 52.6 & 17.5 & 15.0 & 35.0 & 50.0 \\
        MedGemma-27B   & T1CE & 43.5 & 39.1 & 17.4 & 36.2 & 40.6 & 23.2 \\
        MedGemma-27B   & T2   & 43.5 & 37.9 & 18.6 & 45.3 & 29.1 & 25.6 \\
        \bottomrule
    \end{tabularx}
    \caption{The nature of diagnostic flips. Values indicate the proportion of altered predictions that resulted in degradation (Correct$\rightarrow$Wrong), incidental correction (Wrong$\rightarrow$Correct), or abstention ($\rightarrow$UNSURE), aggregated across visual and textual perturbations.}
    \label{tab:flip-directionality}
\end{table*}

\subsection{Illustrative Paired Model Comparisons}
\label{sec:appendix:paired_tests}

On T2 Reverse Order, MedGemma-27B has a 34.4-percentage-point higher Flip Rate than GPT-5.4, with a paired bootstrap 95\% confidence interval of [22.2, 46.7]. An exact McNemar test yields $p=7.84\times10^{-7}$, with 5 GPT-5.4-only flips and 36 MedGemma-27B-only flips.

On T2 Random Shuffle, Qwen3-VL-32B exceeds GPT-5.4 by 16.7 percentage points, with a paired bootstrap 95\% confidence interval of [5.6, 27.8]. These comparisons are computed on the common set of 90 subjects with valid paired outputs.
These selected comparisons illustrate prominent subject-level contrasts and are not intended as an exhaustive ranking of models or perturbations.

\subsection{Granular Breakdown of Diagnostic Flips} 
\label{sec:appendix:extended_results:flips} 

Not all diagnostic fluctuations carry the same clinical implications. To better understand these instabilities, we categorize altered predictions into three directions: degradation (Correct$\rightarrow$Wrong), incidental correction (Wrong$\rightarrow$Correct), and abstention ($\rightarrow$\texttt{UNSURE}). As shown in \cref{tab:flip-directionality}, these aggregated values represent relative proportions conditional on a flip actually occurring. A comprehensive breakdown of these proportions for each specific perturbation is provided in \cref{tab:appendix-detailed-flips}.

Across most models, perturbation-induced changes are not limited to benign uncertainty shifts and often include categorical diagnostic changes. This suggests that presentation changes can alter the diagnostic decision itself, rather than merely increasing uncertainty.
A notable behavioral pattern emerges in GPT-5.4: across the evaluated visual and textual perturbations, its flips occur between categorical diagnoses rather than through a shift to \texttt{UNSURE}. Specifically, none of GPT-5.4's perturbation-induced prediction changes terminate in \texttt{UNSURE}; this 0.0\% value is conditional on a flip and does not denote the model's overall abstention rate. Its high T2 incidental-correction proportion (e.g., 82.9\% under aggregate visual perturbations) is partly influenced by lower baseline accuracy in that modality, which leaves more room for an initially incorrect prediction to change to the correct class. The absence of abstention among flipped cases therefore indicates that output instability can occur without a corresponding shift toward uncertainty.

We also observe that failure modes differ depending on the imaging modality. In T1CE sequences, flips more often manifest as categorical swaps. In contrast, for T2 sequences, models such as Gemini-2.5-Pro, Qwen3-VL-32B, and MedGemma-27B show a greater tendency for flips to result in abstention. For example, under the textual Label Swap perturbation in T2, 67\% of Qwen3-VL-32B's flips result in an \texttt{UNSURE} output. This suggests that, for this cohort and model set, the interaction between modality-specific visual evidence and textual framing can influence whether perturbations produce categorical changes or uncertainty shifts.

Finally, granular inspection reveals distinct sensitivities to evidence positioning. A prominent example is MedGemma-27B's performance under the \textit{ROI Last} configuration in T2 sequences. When the task-critical evidence is shifted to the extreme end of the sequence, 67\% of the model's diagnostic flips result in \texttt{UNSURE}. This pattern suggests that evidence position within the input sequence may affect not only the final diagnosis, but also the model's willingness to abstain.

\subsection{Self-Reported Evidence Localization}
\label{sec:appendix:extended_results:localization}

\Cref{tab:localization-iou} reports the average Intersection over Union (IoU) between the model-reported lesion-visible slice indices and the expert-annotated lesion slices. The table presents these scores for the baseline configuration and all evaluated visual and textual perturbations. These values characterize the stability of the models' explicit localization outputs across presentation changes; however, autoregressive reporting of lesion indices does not establish that the reported slices causally determine the final classification.

\subsection{Lesion Laterality Stability} \label{sec:appendix:extended_results:laterality}

Beyond the primary diagnostic classification, we evaluate the models' spatial reasoning consistency by examining lesion laterality predictions (Left, Right, Bilateral, or Cannot Determine) under a strictly standardized DICOM left-right orientation. \Cref{tab:appendix-laterality} reports the overall rate of laterality changes relative to baseline and decomposes these changes into categorical Left/Right swaps, loss of spatial confidence through a transition to ``Cannot Determine,'' and shifts toward a Bilateral prediction.

Crucially, this analysis measures intra-subject spatial stability rather than baseline laterality correctness. Regardless of whether a model's initial laterality prediction is correct, evidence-preserving sequence reordering should not artifactually trigger a spatial shift. Reversing the sequential order changes the relative position of the images within the model's context window without mirroring the visual content of the individual frames. Consequently, the actual anatomical laterality of the lesion remains invariant.

However, under simple visual reordering (e.g., \textit{Reverse Order}), models such as GPT-5.4 exhibited categorical laterality flips—shifting their prediction from the left to the right hemisphere, or vice versa—in up to 20.5\% of T2 cases. Similarly, under meaning-preserving lexical rephrasing, MedGemma-27B exhibited a pronounced drop in spatial confidence, defaulting to ``Cannot Determine'' in 50.0\% of initially valid T1CE predictions (detailed results are provided in \cref{tab:appendix-laterality}). These findings show that laterality predictions can be sensitive to sequence order even when the intrinsic left-right anatomy of individual images remains unchanged.

\subsection{Blinded Clinical Evidence-Sufficiency Audit}
\label{sec:appendix:clinical_audit}

We conduct a stratified blinded audit of 30 slice-removal sequences. The audited subset contains 15 T1CE and 15 T2 sequences, with 15 GBM and 15 MET cases overall. Subjects are sampled without replacement, and no subject contributes more than one sequence. Case order is randomized, and the reader is not shown subject identifiers, ground-truth labels, the number of removed slices, or the original sequence length.

For each sequence, the neuroradiologist selects one diagnostic response from \texttt{GBM}, \texttt{MET}, and \texttt{UNSURE}. The reader separately records whether the remaining visual evidence is sufficient for a definitive GBM-versus-MET diagnosis.

The neuroradiologist selects \texttt{UNSURE} for all 30 sequences and judges the remaining evidence insufficient in all 30 cases (100.0\%; exact 95\% binomial CI: 88.4--100.0). This specialist assessment supports the expected abstention target used in the model evaluation.

\subsection{Explicit Abstention Guidance}
\label{sec:appendix:safety_calibration}

One alternative explanation for categorical outputs under evidence ablation is that model post-training may discourage abstention even when visual evidence is insufficient. We therefore repeat the GPT-5.4 baseline and slice-removal conditions with explicit guidance to select \texttt{UNSURE} when the available evidence is insufficient for definitive diagnosis.

Baseline abstention remains nearly unchanged, from 3.3\% to 3.3\% on T1CE and from 8.9\% to 10.0\% on T2. Under slice removal, abstention increases from 73.3\% to 80.0\% on T1CE and from 90.0\% to 97.8\% on T2. This selective increase indicates that diagnostic commitment is partly instruction-sensitive rather than reflecting indiscriminate prompt-induced abstention. However, explicit guidance does not fully eliminate categorical prediction, which persists in 20.0\% of the guided T1CE cases.

\subsection{External BraTS Cohort}
\label{sec:appendix:external_cohort}

BraTS provides multi-institutional pre-operative multimodal MRI from high-grade and lower-grade glioma cases, including native T1, post-contrast T1-weighted, T2-weighted, and T2-FLAIR volumes \citep{6975210,bakas2017advancing,bakas2019identifyingbestmachinelearning}. We select 100 subjects with available diagnostic-grade labels, balanced between 50 HGG and 50 LGG cases. For each subject, we use the T1CE and T2 volumes, yielding 200 volumetric sequences.

We preserve the organizer-provided spatial preprocessing and convert each axial volume into an ordered sequence of image slices using the same image-export pipeline applied throughout our evaluation. Subject identifiers and original slice indices are not exposed to the VLM. Each image is presented through the virtual-indexing mechanism described in \cref{sec:methodology:pipeline}.

\begin{table}[t]
    \centering
    \small
    \setlength{\tabcolsep}{4pt}
    \renewcommand{\arraystretch}{1.08}
    \begin{tabularx}{\columnwidth}{ll *{3}{>{\centering\arraybackslash}X}}
        \toprule
        \textbf{Model} & \textbf{Mod.} &
        \textbf{Accuracy} &
        \textbf{Reverse Flip} &
        \textbf{Label Flip} \\
        \midrule
        GPT-5.4 & T1CE & 51.0 & 8.0 & 3.0 \\
        GPT-5.4 & T2   & 59.0 & 14.0 & 12.0 \\
        Qwen3-VL-32B & T1CE & 73.0 & 14.0 & 6.0 \\
        Qwen3-VL-32B & T2   & 66.0 & 26.0 & 15.0 \\
        \bottomrule
    \end{tabularx}
    \caption{Cross-task external evaluation on 100 BraTS subjects under HGG-versus-LGG classification.}
    \label{tab:appendix-external-validation}
\end{table}

\section{Experimental Setup and Reproducibility}
\label{sec:appendix:setup}

\begin{table}[h]
	\centering
	\small
	\setlength{\tabcolsep}{0pt} 
	\renewcommand{\arraystretch}{1.2}
	\begin{tabular*}{\columnwidth}{@{\extracolsep{\fill}} lcc @{}}
		\toprule
		\textbf{Model} & \textbf{Snapshot} & \textbf{Knowledge Cut-off} \\
		\midrule
		Gemini-2.5-Pro & Jun 2025 & Jan 2025 \\
		GPT-5.4        & Mar 2026 & Aug 2025 \\
		\bottomrule
	\end{tabular*}
	\caption{Evaluated proprietary API models, snapshot versions, and knowledge cut-off dates.}
	\label{tab:appendix-models}
\end{table}

\subsection{API Details and Computational Infrastructure}
\label{sec:appendix:api}

To facilitate strict reproducibility, all queries to the evaluated VLMs were executed using a generation temperature of $0.0$, minimizing non-deterministic sampling artifacts. For the proprietary generalist models, the specific API snapshots utilized during our evaluation window are documented in \cref{tab:appendix-models}. Local inference was executed on a system equipped with two NVIDIA RTX 3090 (24GB) GPUs.

\subsection{System Prompts}
\label{sec:appendix:prompt}

The principal system prompts used across experiments are provided below.

\begin{tcolorbox}[
    fonttitle=\footnotesize\bfseries,
    colback=gray!5!white,
    colframe=black!60,
    left=3pt,
    right=3pt,
    top=2pt,
    bottom=2pt,
    title=Standard Baseline Evaluation Prompt,
    fontupper=\fontsize{8pt}{9.5pt}\selectfont,
]
You are a neuroradiologist specializing in neuro-oncology. Your task is to analyze a sequence of axial brain MRI slices and perform a differential diagnosis between Glioblastoma (GBM) and Brain Metastasis (MET) based strictly on the provided visual evidence.

Classify the case as GBM or MET. If the findings are ambiguous or the evidence is insufficient to formulate a diagnosis, you must select UNSURE.

\textbf{Input Format:}\\
You will receive a sequence of images. Each image is immediately preceded by a text tag (e.g., "Image 1:", "Image 2:").

\textbf{Constraints:}\\
- Output MUST be a single, valid JSON object.\\
- Do not include markdown formatting or any preamble.\\
- "Lesion\_Visible\_Indices" MUST be a JSON array of integers corresponding EXACTLY to the numerical indices from the provided "Image X:" text tags where the lesion and its associated peritumoral edema is visible. If no target finding is identified, return an empty array []. 

\textbf{Field Definitions:}\\
- "Primary\_Diagnosis": Exactly one of ["GBM", "MET", "UNSURE"].\\
- "Lesion\_Visible\_Indices": An array of integers (e.g., [14, 15, 16]).\\
- "Lesion\_Laterality": Choose ["Left", "Right", "Bilateral", "Cannot\_Determine"] to indicate the anatomical laterality of the lesion.

\textbf{Output JSON Schema:}
  "Primary\_Diagnosis": "GBM" | "MET" | "UNSURE", \\
  "Lesion\_Visible\_Indices": [integer], \\
  "Lesion\_Laterality": "Left" | "Right" | "Bilateral" | "Cannot\_Determine" 

\label{fig:prompt}
\end{tcolorbox}

\begin{tcolorbox}[
    fonttitle=\footnotesize\bfseries,
    colback=gray!5!white,
    colframe=black!60,
    left=3pt,
    right=3pt,
    top=2pt,
    bottom=2pt,
    title=Textual Perturbation 1: Target Label Swap Prompt,
    fontupper=\fontsize{8pt}{9.5pt}\selectfont,
]
You are a neuroradiologist specializing in neuro-oncology. Your task is to analyze a sequence of axial brain MRI slices and perform a differential diagnosis between Brain Metastasis (MET) and Glioblastoma (GBM) based strictly on the provided visual evidence.

Classify the case as MET or GBM. If the findings are ambiguous or the evidence is insufficient to formulate a diagnosis, you must select UNSURE.

\textbf{Input Format:}\\
You will receive a sequence of images. Each image is immediately preceded by a text tag (e.g., "Image 1:", "Image 2:").

\textbf{Constraints:}\\
- Output MUST be a single, valid JSON object.\\
- Do not include markdown formatting or any preamble.\\
- "Lesion\_Visible\_Indices" MUST be a JSON array of integers corresponding EXACTLY to the numerical indices from the provided "Image X:" text tags where the lesion and its associated peritumoral edema is visible. If no target finding is identified, return an empty array [].

\textbf{Field Definitions:}\\
- "Primary\_Diagnosis": Exactly one of ["MET", "GBM", "UNSURE"].\\
- "Lesion\_Visible\_Indices": An array of integers (e.g., [14, 15, 16]).\\
- "Lesion\_Laterality": Choose ["Left", "Right", "Bilateral", "Cannot\_Determine"] to indicate the anatomical laterality of the lesion.

\textbf{Output JSON Schema:}
  "Primary\_Diagnosis": "MET" | "GBM" | "UNSURE", \\
  "Lesion\_Visible\_Indices": [integer], \\
  "Lesion\_Laterality": "Left" | "Right" | "Bilateral" | "Cannot\_Determine"

\label{fig:prompt_swapped}
\end{tcolorbox}

\vspace{0.3cm}

\begin{tcolorbox}[
    fonttitle=\footnotesize\bfseries,
    colback=gray!5!white,
    colframe=black!60,
    left=3pt,
    right=3pt,
    top=2pt,
    bottom=2pt,
    title=Textual Perturbation 2: Lexical Rephrasing Prompt,
    fontupper=\fontsize{8pt}{9.5pt}\selectfont,
]
Act as an expert in neuro-oncological imaging. Your objective is to evaluate a series of axial brain MRI scans to conduct a differential diagnosis between Glioblastoma (GBM) and Brain Metastasis (MET), relying solely on the visual data provided.

Determine if the pathology is GBM or MET. In cases where the visual signs are unclear or the data is inadequate for a firm conclusion, choose UNSURE.

\textbf{Input Format:}\\
You are presented with a set of images. Every image is labeled with a preceding text tag (for example, "Image 1:", "Image 2:").

\textbf{Constraints:}\\
- The required output is a single, strictly formatted JSON object.\\
- Avoid any introductory remarks or markdown syntax.\\
- The "Lesion\_Visible\_Indices" field must be a JSON array containing the exact integer IDs from the "Image X:" tags where the tumor and related edema are apparent. Return an empty list [] if no abnormality is detected.

\textbf{Field Definitions:}\\
- "Primary\_Diagnosis": Must be one of ["GBM", "MET", "UNSURE"].\\
- "Lesion\_Visible\_Indices": A list of integers (e.g., [14, 15, 16]).\\
- "Lesion\_Laterality": Indicate the position as ["Left", "Right", "Bilateral", "Cannot\_Determine"].

\textbf{Output JSON Schema:}
  "Primary\_Diagnosis": "GBM" | "MET" | "UNSURE", \\
  "Lesion\_Visible\_Indices": [integer], \\
  "Lesion\_Laterality": "Left" | "Right" | "Bilateral" | "Cannot\_Determine"

\label{fig:prompt_rephrased}
\end{tcolorbox}

\subsection{Image Input Protocol}
\label{sec:appendix:limits}

To rigorously test sequential reasoning without introducing synthetic spatial artifacts, we adhered to a strict image formatting protocol:

\paragraph{Independent Frame Processing.}
Each MRI slice was passed to the model as an individual, distinct image within a single multimodal context window. We specifically prohibited image compositing heuristics (e.g., creating grid mosaics), as combining multiple slices into a single file inherently destroys the sequential nature of the data and alters native image resolution.

\paragraph{Model Capacity Constraints.}
The proprietary APIs (GPT-5.4, Gemini-2.5-Pro) and the open-source Qwen3-VL-32B successfully processed all volumetric sequences across their full lengths. In contrast, the domain-specific MedGemma-27B model encountered operational limits on exceptionally long sequences. Specifically, 2 out of the 180 evaluation sequences comprised approximately 160 slices, which exceeded MedGemma-27B's maximum active context length, resulting in incomplete instruction adherence. To maintain absolute experimental control and avoid introducing confounding variables, these two sequences were excluded from the MedGemma-27B evaluation subset, rather than applying an artificial truncation or downsampling policy. Altering the native slice composition for a single model would introduce an unequal data distribution, thereby violating the controlled input-composition requirements of the evaluation.

\end{document}